\documentclass[10pt,twocolumn,letterpaper]{article}

\usepackage{iccv}              

\usepackage{multirow} 

\definecolor{iccvblue}{rgb}{0.21,0.49,0.74}
\usepackage[pagebackref,breaklinks,colorlinks,allcolors=iccvblue]{hyperref}

\def\paperID{9179} 
\def\confName{ICCV}
\def\confYear{2025}

\title{LASER: Tuning-Free LLM-Driven Attention Control for Efficient Text-conditioned Image-to-Animations}

\author{
Haoyu Zheng$^1$ \quad Wenqiao Zhang$^1$ \quad Yaoke Wang$^1$ \quad Juncheng Li$^1$ \quad Zheqi Lv$^1$ \\
Xin Min$^1$ \quad Mengze Li$^1$ \quad Dongping Zhang$^1$ \quad Siliang Tang$^1$ \quad Yueting Zhuang$^1$ \\
\small{$^1$Zhejiang University} \\
\small{$^1$\{\texttt{zhenghaoyu, wenqiaozhang, yaokewang, lijuncheng, lvzheqi,}} \\
\small{\texttt{minxin, limengze, zhangdongping, siliang, yzhuang}\}@zju.edu.cn}
}

\begin{document}
\newcommand*{\method}{LASER}

\twocolumn[{
\renewcommand\twocolumn[1][]{#1}%
\maketitle
\begin{center}
    \vspace{-1em}  
    \centering
    \captionsetup{type=figure}
    \includegraphics[width=\textwidth]{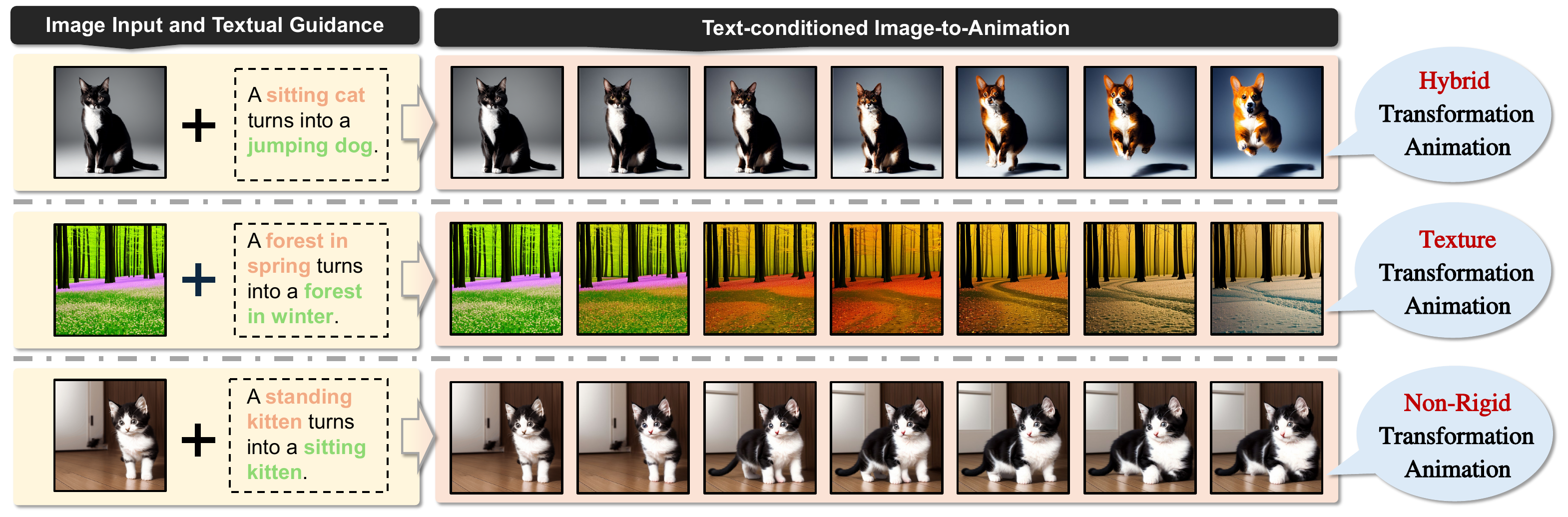}
    \captionof{figure}{Given multimodal inputs (image and textual guidance), our method is capable of guiding the generation of smooth animations based on textual content.}
    \label{fig:teaser}

\end{center}
}]

\begin{abstract}
Revolutionary advancements in text-to-image models have unlocked new dimensions for sophisticated content creation, \emph{e.g.}, text-conditioned image editing, allowing us to edit existing images based on textual guidance. This enables the creation of diverse images that convey highly complex visual concepts.
Despite being promising, existing methods focus on generating new images based on text-image pairs, which struggle to produce fine-grained animations based on existing images and textual guidance without fine-tuning.
In this paper, we introduce a tuning-free LLM-driven attention control framework,  encapsulated by the progressive process of \textbf{L}LM planning $\rightarrow$ feature-\textbf{A}ttention injection $\rightarrow$ \textbf{S}tabl\textbf{E} animation gene\textbf{R}ation, abbreviated as \textbf{\method{}}: 
i) Starting with a general description, \method{} uses a large language model (LLM) to refine it into fine-grained prompts that guide pre-trained text-to-image models in generating aligned keyframes with subtle variations;
ii) The LLM generates control signals for feature and attention injections, enabling seamless text-guided image morphing for various transformations without additional fine-tuning;
iii) Using the same initial noise inversion from the input image, the model receives LLM-controlled injections during denoising and leverages interpolated text embeddings to produce a series of coherent animation frames.
The proposed \method{} introduces a novel tuning-free framework that integrates LLM with pre-trained text-to-image models to facilitate high-quality text-conditioned image-to-animation translation.
In addition, we propose a Text-conditioned Image-to-Animation Benchmark to validate the effectiveness and efficacy of \method{}. Extensive experiments show that the proposed \method{} produces impressive results in both consistent and efficient animation generation, positioning it as a powerful tool for producing detailed animations and opening new avenues in digital content creation.
\end{abstract}    
\section{Introduction}
\label{sec:intro}

Diffusion models~\cite{dhariwal2021diffusion, ho2020denoising, nichol2021improved} form a category of deep generative models that has recently become one of the hottest topics in multimodal intelligence, showcasing impressive capabilities of text-to-image (T2I) generation, ranging from the high level of details to the diversity of the generated examples. Such diffusion models also unlock a new world of creative
processes in content creation, \emph{e.g.}, text-guided image editing \cite{brooks2023instructpix2pix, cao2023masactrl, hertz2022prompt}, involves editing the diverse images that convey highly complex visual concepts with text-to-image models solely through the textual guidance.
Broadly, the contemporary image editing paradigm can be summarized in two aspects: i) \emph{Texture editing}~\cite{brooks2023instructpix2pix, cao2023masactrl, hertz2022prompt}, manipulating a given image's stylization and appearance while maintaining the input structure and scene layout; ii) \emph{Non-rigid Editing}~\cite{cao2023masactrl,kawar2023imagic}, enabling non-rigid image editing (\emph{e.g., } posture changes) while preserving its original characteristics.

Despite achieving impressive image-level editing effects, e.g., creating new images from existing ones, the aforementioned methods fail to generate new animations from existing images according to the user's textual requirements. This includes challenges in handling fine-grained texture and non-rigid transformations. Such text-conditioned image-to-animation serves as an imperative component in various real-world content creation tasks, ranging from cinematic effects to computer games,  as
well as photo-editing tools for artistic and entertainment purposes to enrich people’s imagination. Nevertheless, realizing animation-level generation is highly challenging, primarily due to the highly unstructured latent space of the intermediary images. Of course, we can introduce more animation data to fine-tune the entire T2I diffusion models, thereby capturing the smooth animation generation. However, it comes at a tremendous cost and deteriorates the flexibility of the pre-trained diffusion models under the animation-level generation setting. Based on the above insights, one question is thrown: \texttt{Given the input image and textual description, could we achieve the high-quality animation generation effect with the pre-trained T2I models without fine-tuning?}\!\!\!\!\!\!\!

In this paper, we introduce a novel tuning-free LLM-driven attention control framework for text-conditioned image-to-animation, through  \textbf{L}LM planing $\rightarrow$ feature-\textbf{A}ttention injection $\rightarrow$ \textbf{S}tabl\textbf{E} animation gene\textbf{R}ation, named as \textbf{\method{}}. The core of our framework is that by leveraging the large language models (LLMs)\cite{touvron2023llama, achiam2023gpt, li2023fine, zhang2024hyperllava} with significant potential in natural language processing, to effectively parse the input textual description into relevant and continuous control statements for pre-trained T2I diffusion models, thereby transforming the given image to animation. Specifically, \method{} comprises the following progressive steps: \textbf{Step 1}, given a multimodal input, \emph{i.e.}, a description of the animation \(P_*\) and an initial image \(I_0\) (which can be optional, allowing the T2I model generation), LLM decomposes the general and coarse-grained description \(P_*\) into multiple fine-grained and consistent prompts. These prompts are closely aligned and exhibit subtle variations, aiding in the guided editing of subsequently corresponding keyframes.
\textbf{Step 2}, the LLM converts these prompts into control signals for feature and attention injections, tailored to the subtle differences between adjacent prompts. This allows for customized injection strategies based on three categories: Feature and Association Injection (FAI) for texture transformations, Key-Value Attention Injection (KVAI) for non-rigid transformations, and Decremental Key-Value Attention Injection (DAI) for hybrid transformations. The model performs DDIM Inversion on the input image to obtain initial noise, extracting features and self-attention values in the process.
\textbf{Step 3}, for each animation frame, the model begins by copying the initial noise obtained from the DDIM Inversion of the input image or previous frame. The frame's text embedding is then derived by interpolating the embeddings of adjacent text prompts. Throughout the denoising process, the model receives LLM-controlled feature and attention injections. This method ensures the generation of a sequence of coherent and visually aligned animation frames.

Additionally, we inaugurate a Text-conditioned Image-to-Animation Benchmark, a comprehensive collection designed to challenge and quantify the adaptability and precision of the proposed \method{}.

Summing up, our contributions can be concluded as:
\begin{itemize}
    \item We introduce the tuning-free text-conditioned image-to-animation task based on the T2I model, designed to craft high-quality animations based on the multimodal input using the pre-trained text-to-image models, without additional fine-tuning or annotations.
    To evaluate the efficacy of our approach, we introduce the \textit{Text-conditioned Image-to-Animation Benchmark}, hoping that it may support future studies within this domain.
    \item The proposed \method{} encapsulated by the progressive process of LLM planing $\rightarrow$ Feature-attention injection $\rightarrow$ Stable animation generation, enabling the smooth texture and non-rigid transformation animation generation.
    \item Both qualitative and quantitative assessments underscore the superior efficacy of the proposed framework, showcasing its proficiency in generating animations that are not only smooth and of high quality but also diverse. 
\end{itemize}

\section{Related Work}
\label{sec:related_work}
\noindent\textbf{Text-to-Image Generation.}
In artificial intelligence\cite{zhang2022boostmis,zhang2019frame,zhang2023revisiting}, text-to-image (T2I) generation focuses on creating high-quality images from text descriptions. Earlier approaches, primarily based on Generative Adversarial Networks (GANs) \cite{brock2018large, zhu2019dm}, aligned text with synthesized images using multimodal vision-language learning, yielding impressive results on specific datasets. Recently, diffusion models \cite{ho2020denoising, nichol2021improved, dhariwal2021diffusion} have surpassed GANs, offering state-of-the-art generation quality and diversity. Text-to-image diffusion models \cite{rombach2022high, saharia2022photorealistic}, conditioned on text via cross-attention layers, ensure that generated images are both visually and semantically aligned with the input descriptions.

\noindent\textbf{Text-guided Image Editing.}
Text-guided image editing modifies images using textual descriptions, enabling users to specify changes in natural language. While GAN-based methods \cite{li2020manigan, patashnik2021styleclip, xia2021tedigan} have shown success, they often lack generalizability. VQGANCLIP \cite{crowson2022vqgan} combines VQGAN \cite{esser2021taming} and CLIP \cite{radford2021learning} for high-quality edits but suffers from slow generation and high computational costs. Recent diffusion models like Imagen \cite{saharia2022photorealistic} and Stable Diffusion \cite{rombach2022high} excel in text-to-image generation and serve as robust priors for text-guided image manipulation \cite{cao2023masactrl, hertz2022prompt, kawar2023imagic, tumanyan2023plug}. Techniques like Prompt-to-Prompt \cite{hertz2022prompt} and Plug-and-Play \cite{tumanyan2023plug} use cross-attention or spatial features for editing, while MasaCtrl \cite{cao2023masactrl} and Imagic \cite{kawar2023imagic} handle non-rigid transformations.

\noindent\textbf{Image Morphing.}
Image morphing in computer graphics and image processing involves creating smooth transitions between two images by generating intermediate frames \cite{aloraibi2023image, zope2017survey}. Deep learning techniques, including neural networks, have been employed to identify correspondences and generate intermediate images through latent interpolations. GAN-based methods \cite{karras2021alias, karras2019style, karras2020analyzing, sauer2023stylegan} have demonstrated impressive morphing results via linear interpolation in latent space. Recent research on diffusion models has explored generating intermediate images through latent noise and text embedding interpolation \cite{bao2023one, song2020score, wang2023interpolating}. Impus \cite{yang2023impus} applied diffusion models for interpolation in both the locally linear continuous text embedding space and Gaussian latent space. DiffMorpher \cite{zhang2023diffmorpher} enhances image morphing by combining spherical linear interpolation on latent noise obtained via DDIM inversion with text-conditioned linear interpolation, effectively addressing challenges in the unstructured latent space of diffusion models.

\section{Methodology}
\label{sec:method}

Given a user-defined descriptor \( P_* \) and an initial image \( I_0 \), our method generates an animation sequence \(\{x_0^{(\alpha)}, x_1^{(\alpha)}, \ldots, x_n^{(\alpha)}\}\), where \(\alpha\) varies from 0 to 1. The length of the \( x_i^{(\alpha)} \) sequence is set by \( n_f \) and the number of sequences \( x_i^{(\alpha)} \) corresponds to the transformation stages \( n_t \). The resulting animation should retain the structure of $I_0$ while incorporating the characteristics described by \( P_* \).

\begin{figure*}[h]
\centering
\includegraphics[width=\linewidth]{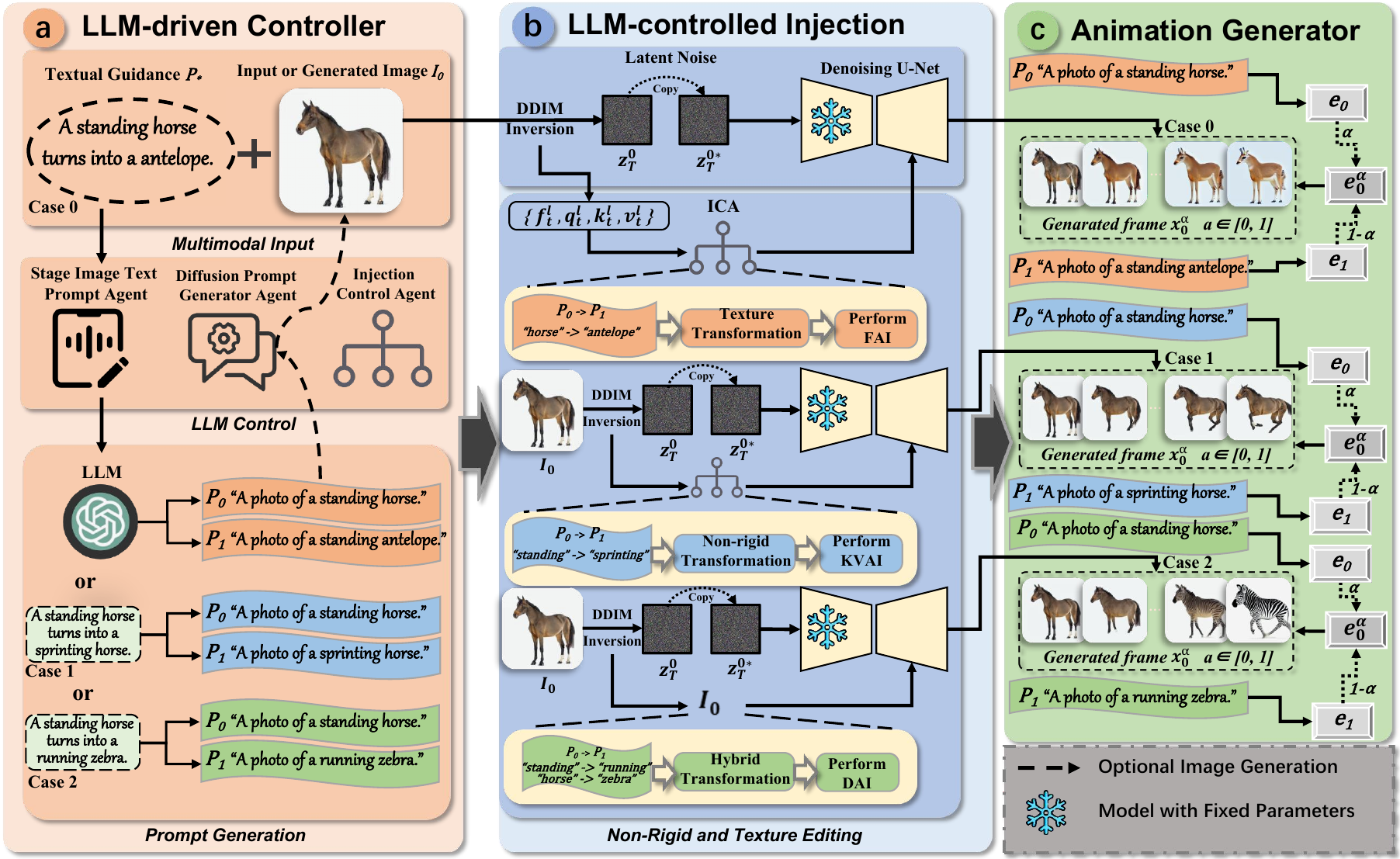}
\caption{Overview of proposed \method{}. (a) The LLM-driven Controller employs three specialized agents: SIA decomposes the animation description into key stage prompts, PGA enhances prompts for initial image generation when needed, and ICA determines appropriate injection strategies based on transformation types. (b) The selected injection strategy (FAI/KVAI/DAI) guides the feature and attention manipulation process for texture and non-rigid editing. (c) The animation generator uses text embedding interpolation to ensure smooth transitions between keyframes, enabling fluid animation.}
\label{fig:method}
\end{figure*}

\subsection{Preliminary for Diffusion Models}

Diffusion models~\cite{ho2020denoising, song2020denoising, nichol2021improved} are probabilistic generative models that generate images by gradual denoising a noise distribution, typically Gaussian noise. The process involves two phases: a \textit{forward} process that adds noise to data, and a \textit{reverse} process that removes noise to reconstruct the image. Starting from pure noise, the reverse process iteratively refines the image by predicting and subtracting noise, ultimately producing a clean image.

We use a text-conditioned Stable Diffusion (SD) model~\cite{rombach2022high} that operates in a lower-dimensional latent space instead of pixel space. Images are first encoded into latent representations by a variational auto-encoder (VAE)~\cite{kingma2013auto_dis5}, followed by diffusion-denoising within the latent space. The denoised latent representation is then decoded back into the image space to produce the final image.

In the noise-predicting network $\epsilon_{\theta}$, residual blocks generate intermediate features $f_t^{l}$, which are used in the self-attention module to produce $Q$, $K$, and $V$ for capturing long-range interactions. Cross-attention then integrates the textual prompt $P$, merging text and image semantics. The attention mechanism is formulated as:
\begin{equation}
    Attention(Q, K, V) = softmax(\frac{QK^T}{\sqrt{d_k}})V
\end{equation}
where $Q$, $K$, and $V$ represent queries, keys, and values, with $d_k$ denoting the key/query dimension. Attention layers in the SD model significantly influence image composition and synthesis, guiding editing by manipulating attention during denoising~\cite{hertz2022prompt, tumanyan2023plug}.

\subsection{LLM-driven Controller} \label{prompt extractor}

To produce high-quality animations guided by textual descriptions, we design an LLM-driven control framework comprising three specialized agents. This framework receives a user-defined descriptor $P_*$ and an optional initial image $I_0$, producing control signals that guide the animation generation process.

\subsubsection{LLM Agent Architecture}

Our framework consists of three LLM agents in a sequential pipeline. The Stage Image Text Prompt Agent (SIA) analyzes the animation description $P_*$ and decomposes it into a series of fine-grained prompts $\{P_0, P_1, \ldots, P_{n_t}\}$ representing key animation stages. SIA follows two core principles: decomposing the animation descriptor into independent processes to reduce semantic differences between prompts, and maintaining consistent sentence structures to ensure smooth transitions in the CLIP embedding space~\cite{kawar2023imagic}. For instance, ``A year has passed on the spring meadow'' would be divided into four seasonal prompts: ``The meadow in spring'', ``The meadow in summer'', ``The meadow in autumn'', and ``The meadow in winter''.

When a user does not provide an initial image, the Stable Diffusion Prompt Generator Agent (PGA) enhances the initial prompt $P_0$ by incorporating elements like texture, lighting, and artistic style, guiding Stable Diffusion in generating a high-fidelity initial image $I_0$.

The Injection Control Agent (ICA) determines the appropriate injection strategy for each transition between animation stages. ICA analyzes prompt pairs $\{P_i, P_{i+1}\}$ and selects one of three strategies: ``Feature and Association Injection'' (FAI) for texture transformations, ``Key-Value Attention Injection'' (KVAI) for non-rigid transformations, or ``Decremental Key-Value Attention Injection'' (DAI) for hybrid transformations.

\subsubsection{Agent Workflow and Decision Process}

As illustrated in Fig.~\ref{fig:method}(a), the agents operate in a coordinated workflow. SIA first processes the animation description to generate key stage prompts. If no initial image is provided, PGA enhances $P_0$ to guide image generation; otherwise, the user-provided image is used directly.

For each transition between stages, ICA analyzes the semantic differences between consecutive prompts. When changes primarily involve appearance attributes such as color or texture, ICA triggers FAI. For pose or major shape modifications, KVAI is selected. For complex transitions requiring both appearance and structural changes, DAI is employed to ensure coherent integration. Control signals from ICA are passed to the injection module described in Section~\ref{sec:injection}, which applies the selected strategy to guide animation frame generation through the process in Section~\ref{sec:generator}. All three agents use the same LLM with different system prompts, requiring no additional fine-tuning, enabling flexible adaptation to various animation types.




\subsection{LLM-controlled Injection} \label{sec:injection}
\begin{figure}[t]
\centering
\includegraphics[width=0.98\linewidth]{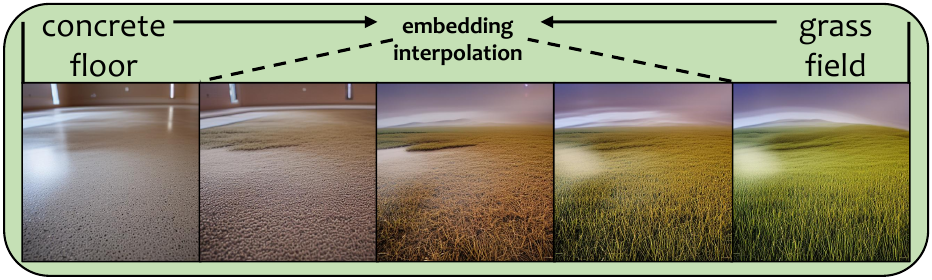}
\caption{By leveraging interpolation of text embeddings, meaningful intermediate images can be generated. }
\label{fig:inter}
\end{figure}

After receiving text prompts \(\{P_0, P_1, \ldots, P_{n_{t}}\}\), the model will sequentially generate the animation sequence \(\{x_0^{(\alpha)}, x_1^{(\alpha)}, \ldots, x_n^{(\alpha)}\}\), where $n$ = $n_t-1$. The animation $x_i^{(\alpha)}$ will depict the transition from $P_i$ to $P_{i+1}$. To maintain a certain structure in the images, the initial noise for each animation frame is obtained through performing DDIM Inversion on the last frame $x^1_{i-1}$ of the previous animation (with the first animation using the input image $I_0$). For a single animation frame $x_i^{\alpha}$, its initial noise $z_T^{i*}$ is copied from $z_T^i$. We then perform a simple linear interpolation between the text embeddings of two adjacent key animation stages to obtain the corresponding text embedding:

\begin{equation}
e_{i}^{\alpha} = (1 - \alpha)\cdot e_i + \alpha \cdot e_{i+1}
\end{equation}
where $e^i$ and $e^{i+1}$ are the text embeddings corresponding to $P_i$ and $P_{i+1}$, respectively. The interpolation parameter $\alpha$ is discretized into a series of equidistant values within the interval $[0, 1]$ to facilitate a meaningful transition between frames. The number of discrete values corresponds to the intended number of frames, denoted as $n_f$. Thus, $\alpha$ takes on values $\alpha_0, \alpha_1, \ldots, \alpha_{n_f-1}$, where $\alpha_0 = 0$ represents the starting frame and $\alpha_{n_f-1} = 1$ indicates the ending frame. This linear spacing ensures a continuous transformation across the animation sequence as shown in Fig.~\ref{fig:inter}.

\begin{figure}[t]
\centering
\includegraphics[width=0.98\linewidth]{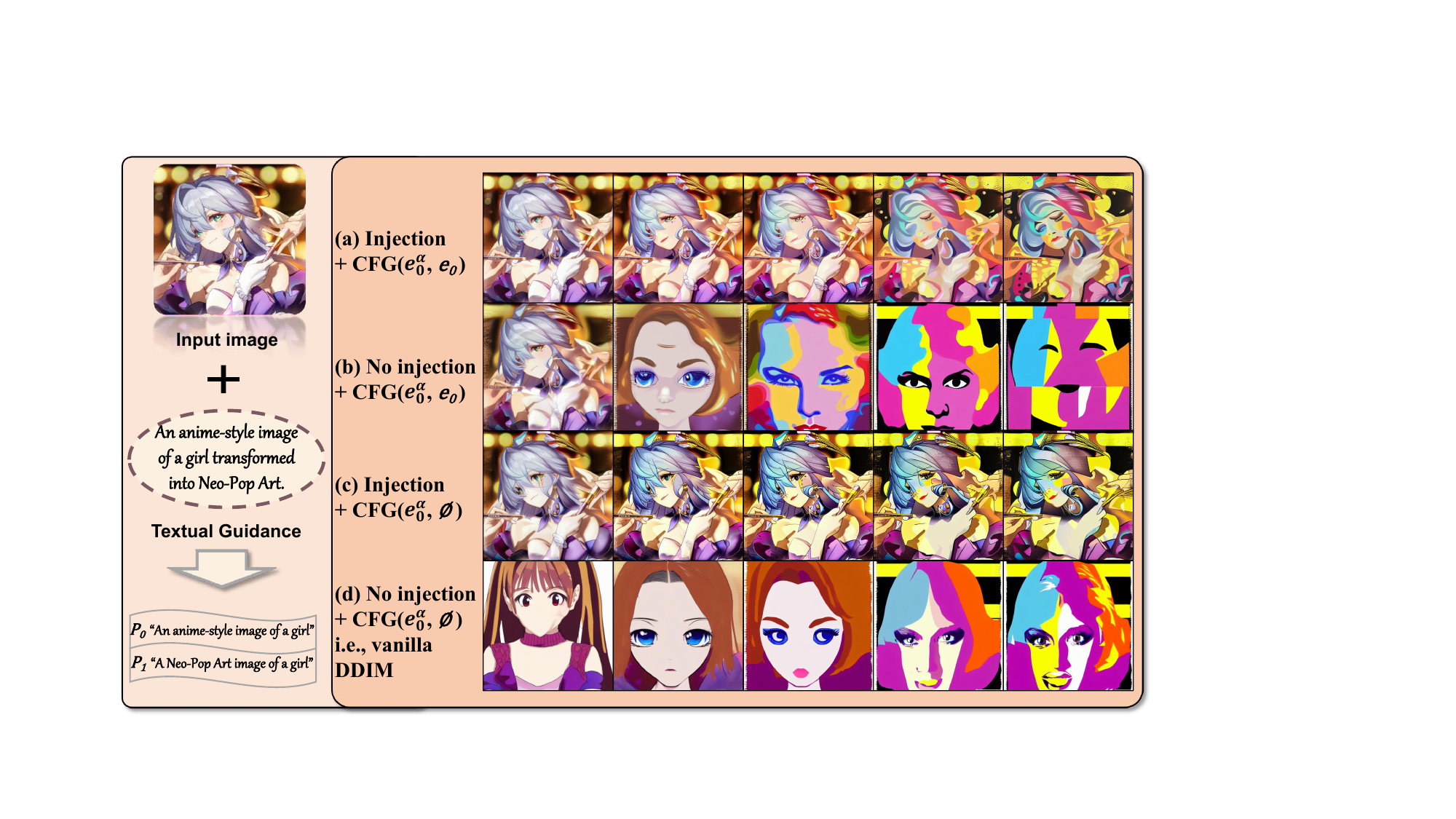}
\caption{Overview of Feature and Association Injection. }
\label{fig:ablation_method}
\end{figure}

However, directly applying the vanilla DDIM process fails to produce desired animation, particularly when the input image is of anime style, a domain at which Stable Diffusion is less proficient. As shown in Fig. \ref{fig:ablation_method} (d), this approach not only fails to align the first frame with the input image but also loses the structural integrity of the input image in subsequent frames, leading to an unsatisfactory result.

\subsubsection{Texture Transformation.}

\begin{figure}[t]
\centering
\includegraphics[width=0.98\linewidth]{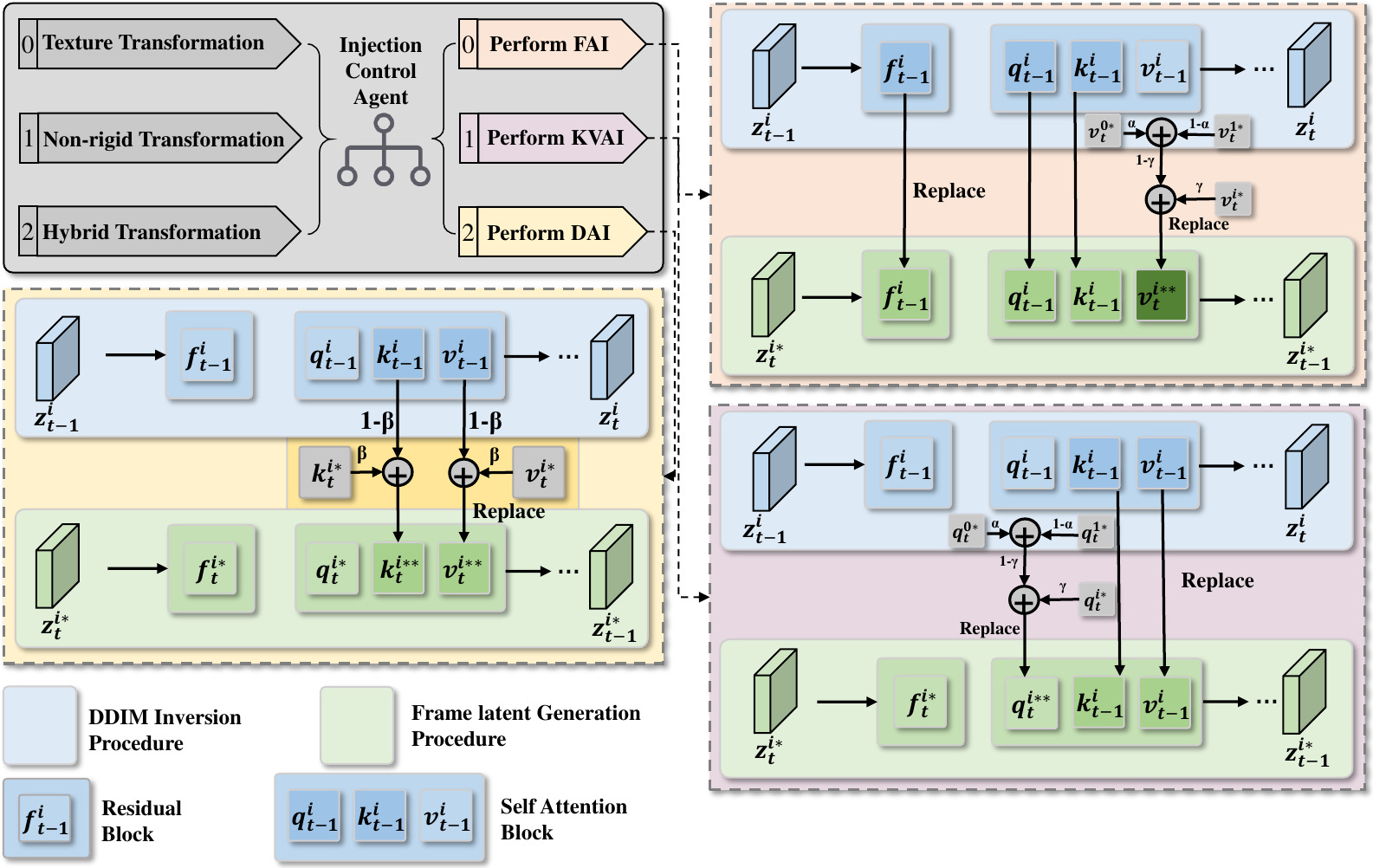}
\caption{Overview of our injection methods. }
\label{fig:Injection}
\end{figure}

To address texture changes, we propose an LLM-controlled feature and attention injection approach. Specifically, we first perform DDIM Inversion on the prior image $I_0$ or $x_{i-1}^1$ to extract residual block features $f_t^i$ and self-attention projections $q_t^i$, $k_t^i$, and $v_t^i$, ultimately obtaining the initial latent representation $z_T^i$. Our method, termed "Feature and Association Injection" (FAI), injects selected features $f_{t-1}^i$ into the fourth layer of residual blocks, while propagating self-attention elements $q_{t-1}^i$ and $k_{t-1}^i$ across decoder layers, as illustrated in Fig.~\ref{fig:Injection}.
During the denoising process at each frame, we perform a two-step interpolation on the value components. First, we interpolate between the values from the first and last frames based on the current frame index weight $a_i$. Then, we further blend this intermediate result with the current value component using a time-step decay weight $\lambda = \frac{t}{T}$, formulated as:

\begin{equation}
v_{t}^{i**} = (1-\gamma) \cdot (\alpha \cdot v_{t}^{0*} + (1-\alpha) \cdot v_{t}^{1*}) + \gamma \cdot v_{t}^{i*}  
\end{equation}

The extracted $q$, $k$, and $f$ from the source image effectively preserve structural consistency across animation frames~\cite{tumanyan2023plug}, while the progressive interpolation of $v$ ensures smooth texture transitions, as depicted in Fig.~\ref{fig:ablation_method}~(c).

\subsubsection{Enhancing Generation Quality}

With FAI, animation frame quality improves significantly over vanilla DDIM, but issues remain: (i) the first frame misaligns with the input image, (ii) frames lose detail, and (iii) weak prompt adherence. While classifier-free guidance (CFG) enhances text conditioning, it struggles with guidance scales over 1.0. To address this, we replace the null-text embedding with the input text prompt embedding $e_i$ for frame $x_i^{\alpha}$, as shown in Fig.~\ref{fig:ablation_method}(b). This preserves the input image’s structure while enabling prompt-driven edits~\cite{miyake2023negativepromptinversionfastimage}, resulting in detailed, coherent frames that align with both the input image and textual guidance, as shown in Fig.~\ref{fig:ablation_method}(a).

\subsubsection{Non-Rigid Transformation.} 

While effective, the previous method struggles with non-rigid modifications. To address this, we avoid injecting into residual blocks to preserve image structure and instead apply targeted attention injections. Since image layout stabilizes early in denoising and self-attention queries align semantically~\cite{cao2023masactrl}, they can effectively extract content from various objects. At a specific timestep, we inject keys $k_{t-1}^i$ and values $v_{t-1}^i$ from the inversion stage.
To ensure smooth structural transitions, we linearly interpolate the query between the first and last frames using the current frame weight $a_i$ and further blend it with the corresponding value via time-step decay interpolation, formulated as:

\begin{equation}
k_{t}^{i**} = (1-\gamma) \cdot (\alpha \cdot k_{t}^{0*} + (1-\alpha) \cdot k_{t}^{1*}) + \gamma \cdot k_{t}^{i*}  
\end{equation}

Additionally, we incorporate cross-frame attention during injection, enabling each frame’s query $q$ to access all frames’ keys and values. This enhances structural consistency, refines object outlines from text prompts, and enriches frames with source image details, facilitating semantically coherent non-rigid transitions. We refer to this approach as **Key-Value Attention Injection (KVAI)**, as illustrated in Fig.~\ref{fig:Injection}.

\subsubsection{Hybrid Transformation.}
Additionally, we introduce **Decremental Key-Value Attention Injection** (DAI) to generate animation sequences that integrate both texture changes and non-rigid transformations, as illustrated in Fig.~\ref{fig:Injection}. In this approach, the animation undergoes a complete non-rigid transformation from start to end, while textures evolve progressively under prompt guidance, ultimately retaining only partial features from the input image. Similar to KVAI, DAI follows a comparable processing framework, with the key distinction that it interpolates between $k_{t-1}^i, v_{t-1}^i$ and $k_{t}^{i*}, v_{t}^{i*}$ to obtain $k_{t}^{i**}, v_{t}^{i**}$. This interpolation enables smoother transitions between frames, ensuring coherent integration of both texture and structural transformations. Furthermore, we incorporate **cross-frame attention** during injection to enhance frame consistency, further improving the overall temporal coherence of the animation.

\begin{equation}
k_{t}^{i**} = (1 - \beta)k_{t-1}^i + \beta k_{t}^{i*}, \\
v_{t}^{i**} = (1 - \beta) v_{t-1}^i + \beta v_{t}^{i*},
\end{equation}
where \(\beta = w * \alpha\), \(w \in (0, 1)\). In text embedding interpolation, \(\beta\) can replace \(\alpha\) to achieve a higher-quality final frame. This mechanism ensures a gradual and smooth transition in the hybrid transformation process. The layers and timesteps utilized in DAI are the same as those in KVAI.

\subsection{Animation Generator} \label{sec:generator}
Based on the injection strategy selected by ICA and the corresponding control signals, we create the sequence $x_i^{(\alpha)}$ comprising a series of animation frames $x_i^{\alpha}$. We start by deriving the text embeddings $e_{\alpha}$ through interpolation between $e_i$ and $e_{i+1}$.
Subsequently, the latent variable $z_T^{i*}$ undergoes denoising, during this phase, it is subjected to feature and attention injections governed by the ICA, ultimately producing 
$x_i^{\alpha}$. 
The first frame, denoted by  $\alpha = 0$, precisely reconstructs the input image.
As $\alpha$ progresses from 0 to 1, the influence of $P_i$ gradually decreases,  whereas the influence of $P_{i+1}$ correspondingly escalates.
The sequence $x_i^{(\alpha)}$ is completed with the generation of the final frame $x_i^1$.
If $P_{i+2}$ is available, the model performs DDIM Inversion on $x_i^1$ to obtain a new initial noise $z^{i+1}_T$, and the process is repeated.

\section{Experiments}
\label{sec:experiments}
\subsection{Benchmark and Setting}
\noindent\textbf{Experimental Setting.} 
We employ the publicly available Stable Diffusion v1.5 \cite{rombach2022high} as our diffusion model and use GPT-4 8k \cite{openai2024gpt4} as the LLM. We perform 50 forward steps of DDIM Inversion on the input image to obtain the initial noise, and 50 backward steps of DDIM Sampling to obtain an animation frame. When implementing FAI, we inject features and self-attention within the first 25 of the 50-step sampling process, targeting layers all of the U-Net decoder. In the case of KVAI and DAI, injections commence after the initial five sampling steps and are applied within layers 3 to 8 of the decoder. For sampling, the classifier-free guidance scale is set at 7.5. Runtime evaluations are performed on an NVIDIA RTX 4090 GPU.

\noindent\textbf{Text-conditioned Image-to-Animation Benchmark.} 
Due to the lack of benchmarks for such configurations, we have proposed a new dataset: Text-conditioned Image-to-Animation Benchmark, a dataset with 200 sets of textual descriptions for animation generation. It includes various transformations like animals, objects, natural changes, art styles, and human transformations. The dataset consists of 70 sets with material transformations, 70 with non-rigid transformations, and 60 with hybrid transformations, providing a broad basis for evaluating our model's performance.
To construct the dataset, we used the SIA for key animation stage prompts and the PGA for detailed prompts, which, combined with Stable Diffusion, generated initial images. These images, paired with textual prompts, enabled our model to create 200 animation sequences.

\subsection{Qualitative Evaluation}

We present a visual comparison against prior approaches to demonstrate our method's superiority. While no other tuning-free methods exist for text-controlled image-to-animation based on T2I models, we compare our approach with state-of-the-art baselines in text-controlled image morphing and editing. These include: 1) Diffusion-based morphing methods like DDIM \cite{song2020denoising}, Diff.Interp \cite{wang2023interpolating}, and DiffMorpher \cite{zhang2023diffmorpher}; and 2) Text-driven, tuning-free editing techniques like PnP \cite{tumanyan2023plug} and MasaCtrl \cite{cao2023masactrl}. Details on baseline adaptations for animation generation are in the Appendix.

\begin{figure*}[h]
\centering
\includegraphics[width=\linewidth]{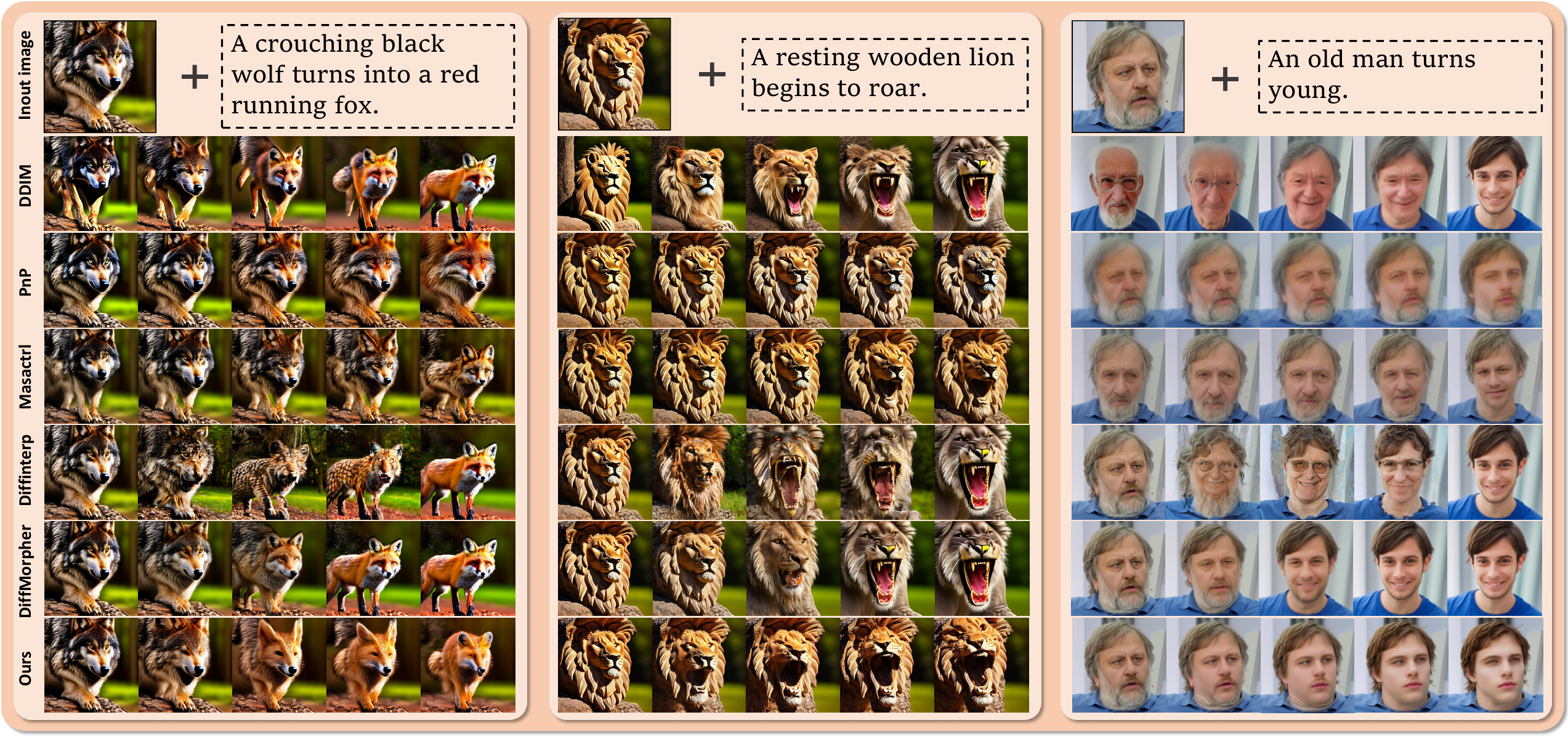}
\caption{Qualitative evaluation. Our method produces animations that significantly outperform previous methods in terms of quality, smoothness, and alignment with user input.}
\label{fig:cmp}
\end{figure*}

\paragraph{\textbf{Generation Results.}}
As shown in Fig. \ref{fig:cmp}, our method outperforms previous approaches in aligning with user input, ensuring smooth transitions, and maintaining both semantic coherence and the subject's identity throughout the animation. Tuning-free methods often fail to align with the input image, while those requiring fine-tuning may initially succeed but gradually lose structure as the animation progresses. When generating hybrid transformations, these methods struggle with expected motion and appearance changes. In contrast, our method excels in producing dynamic non-rigid transformations while preserving object structure. Other methods often lose structural and texture details over time, especially during texture-only changes, whereas ours consistently maintains the character's appearance and produces smooth transitions. While PnP better preserves the structure, it often introduces color inconsistencies and blurriness. Compared to previous methods, our approach consistently generates coherent animations that align closely with user input, yielding visually satisfying results. Leveraging the LLM's prior knowledge, it interprets the same input into diverse textual variations, enabling the LASER model to generate various animations from the same text. Additional examples are in the Appendix.



\subsection{Quantitative Evaluation}
We quantitatively evaluate models using metrics from 12-frame animations, with Runtime measuring the time required to generate 16 frames. Additionally, we compare the runtime of FAI (left) with that of KVAI and DAI (right).

\paragraph{\textbf{Evaluation Metrics.}}
We evaluate our method using complementary metrics that assess different aspects of animation quality. Perceptual Input Conformity (PIC, $\uparrow$)~\cite{xing2023dynamicrafter} measures visual alignment between the generated video and input image, where higher values indicate better preservation of input content. CLIP Score ($\uparrow$)~\cite{radford2021learning} evaluates semantic consistency between frames and their corresponding textual descriptions, with higher values indicating stronger text-image coherence. Learned Perceptual Image Patch Similarity (LPIPS, $\downarrow$)~\cite{zhang2018unreasonable} quantifies perceptual differences within the animation sequence, where we report total perceptual deviation (LPIPS$_T$) and maximum deviation from the nearest endpoint (LPIPS$_M$). Lower values indicate better visual consistency. Perceptual Path Length (PPL, $\downarrow$)~\cite{karras2020analyzing} measures frame-to-frame perceptual variation, where lower values correspond to smoother, more natural transitions. Comprehensive evaluation results are presented in Table~\ref{tab:method_comparison}.

Our method achieves the highest PIC, demonstrating superior visual alignment with the input image. In terms of animation smoothness, our significantly lower LPIPS and PPL indicate more uniform frame transitions. While our CLIP Score is slightly lower than DiffMorpher, which leverages LoRA-based interpolation to generate visually coherent intermediate frames, it suffers from structural instability during deformation, leading to a lower PIC. Fixed-attention injection methods like PnP and Masactrl struggle with diverse animation generation, resulting in less stable frames. In contrast, our LLM-controlled dynamic injections strategies enable flexible adaptation to various transformations, excelling across all metrics. Moreover, our approach operates efficiently without fine-tuning or DDIM reconstruction, offering a significant computational advantage.

\begin{table}[h]
\centering
\captionsetup{font={small,stretch=1}, labelfont={bf}}
\caption{
Comparison of current methods. Superscript \textsuperscript{$\clubsuit$} indicates models that use an external network (e.g., ControlNet~\cite{zhang2023adding}) for generating intermediate images, while \textsuperscript{\dag} denotes models fine-tuned with training LoRA ~\cite{hu2021lora} on images. ``TT'' refers to texture transformation, ``NRT'' to non-rigid transformation, and ``AG'' to animation generation, which produces animation frames based on existing images.
}
\resizebox{1\columnwidth}{!}{
\renewcommand{\arraystretch}{1.25}
\begin{tabular}{l||ccc||ccccc||c}
\hline
\multirow{2}{*}{\textbf{Method}}  & 
\multicolumn{3}{c||}{Characteristics} & \multicolumn{5}{c||}{Metrics} & \multirow{2}{*}{\textbf{Runtime}} \\
\cline{2-4} \cline{5-9}
& TT & NRT & AG & PIC & LPIPS$_T$ & LPIPS$_M$ & CLIP Score & PPL & \\
\hline
DDIM  & \checkmark & & & 0.806 & 1.353 & 0.253 & 0.968 & 14.879 & 34s \\
PnP  & \checkmark & & & 0.865 & 0.974 & 0.183 & 0.978 & 10.718 & 1min3s \\
MasaCtrl  & & \checkmark & & 0.839 & 1.158 & 0.215 & 0.971 & 12.737 & 1min1s \\
Diff.Interp\textsuperscript{$\clubsuit$}  & & & \checkmark & 0.787 & 2.170 & 0.413 & 0.930 & 32.819 & 2min6s \\
DiffMorpher\textsuperscript{\dag}  & & & \checkmark & 0.835 & 0.530 & 0.249 & \textbf{0.986} & 5.826 & 1min46s \\
Ours & \checkmark & \checkmark & \checkmark & \textbf{0.915} & \textbf{0.489} & \textbf{0.076} & 0.984 & \textbf{5.380} & 41s\text{/}48s \\
\hline
\end{tabular}%
}
\label{tab:method_comparison}
\end{table}

\subsection{Ablation Study}
\begin{figure}[ht]
\centering
\includegraphics[width=0.98\linewidth]{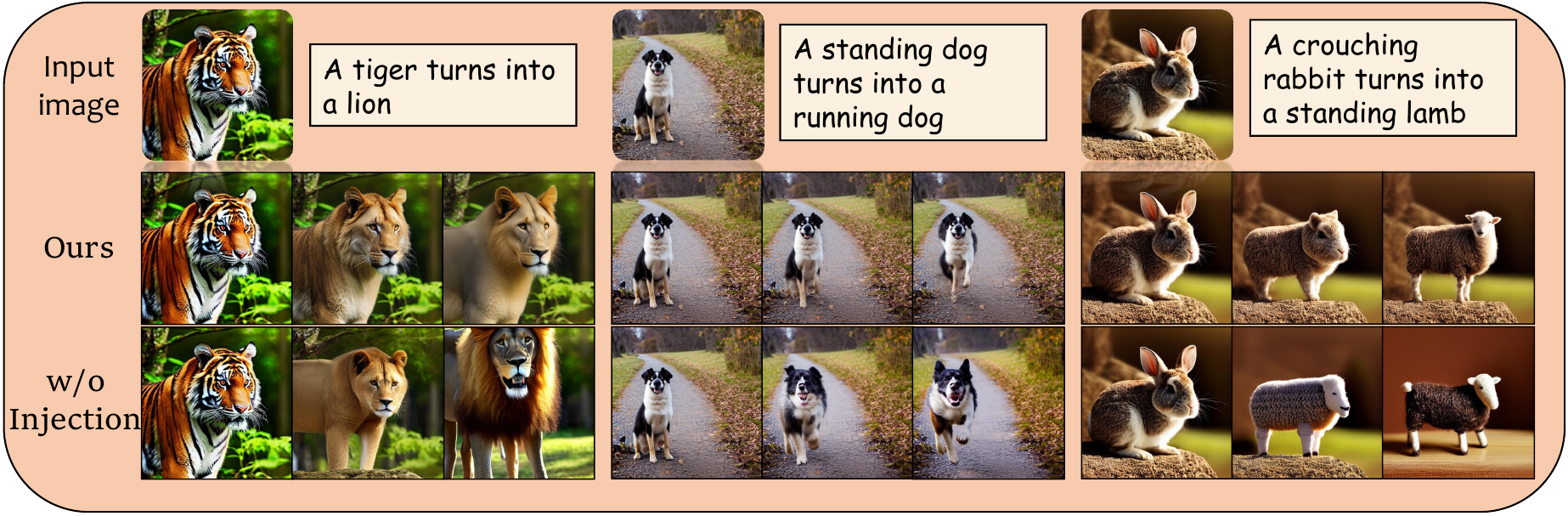}
\caption{Ablation study of three injection methods. Our proposed injection strategies significantly enhance the generation quality across various types of animations.}
\label{fig:ablation_exp}
\end{figure}

To evaluate the effectiveness of the proposed components, we have conducted an ablation study by systematically removing each of the key feature and attention injection mechanisms. As shown in Table~\ref{tab:ablation} and Table~\ref{ablation_2}, the proposed feature and attention injection strategies effectively enhance animation smoothness and maintain consistency with the input images, leading to substantial improvements across various evaluation metrics. The qualitative ablation experiments in Fig.~\ref{fig:ablation_exp} further illustrate the impact of these components. It can be observed that FAI effectively preserves the structural content of the main subject during smooth texture transitions, ensuring that material transformations occur seamlessly without distorting the overall structure. KVAI plays a crucial role in maintaining the stability of the subject's appearance during non-grid transformations, preventing structural inconsistencies when handling non-rigid changes. Meanwhile, DAI ensures smooth transitions between frames in hybrid transformations, preventing abrupt jumps in texture and structure while preserving motion coherence. When ICA is removed, DAI is used by default; in this case, the mismatched injection approach compromises both animation smoothness and fidelity to the input image. The results clearly demonstrate that each injection method and appropriate injection strategy contributes significantly to the overall animation quality, and removing any of them leads to noticeable performance degradation, particularly in complex transformations requiring both texture and structural changes.

\begin{figure}[t]
\centering
\includegraphics[width=0.98\linewidth]{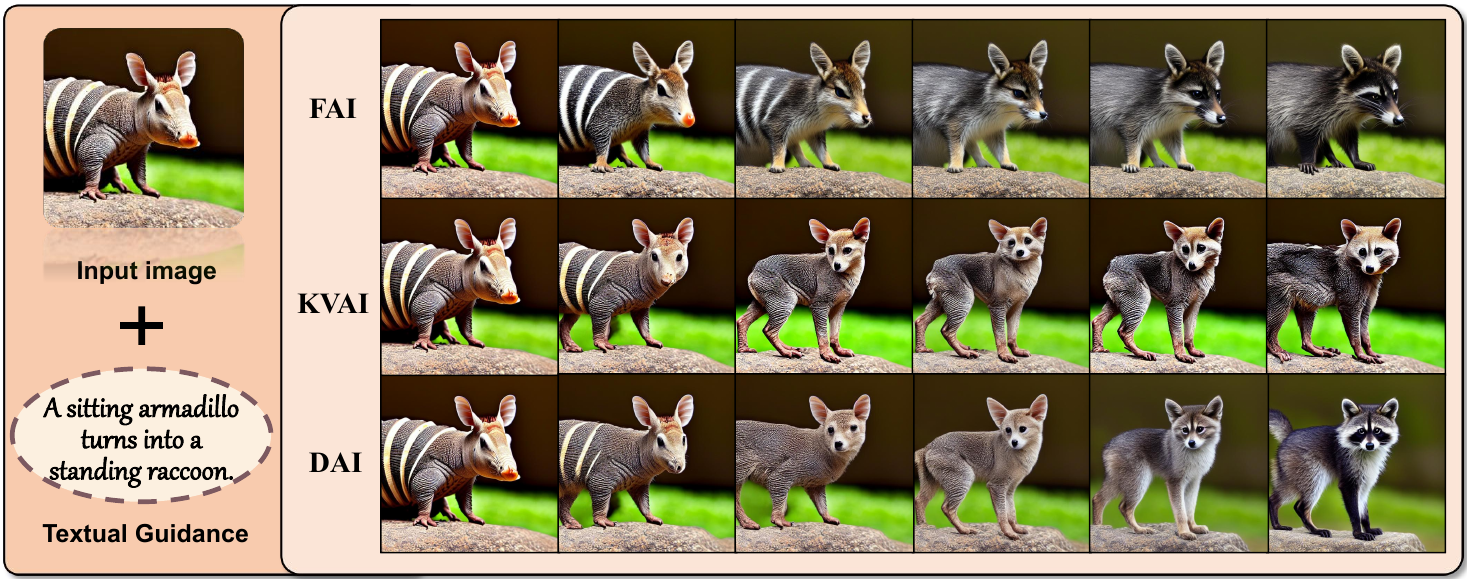}
\caption{Effectiveness of different injection strategies.}
\label{fig:DAI}
\end{figure}

\paragraph{\textbf{Effectiveness of DAI.}}
To demonstrate the effectiveness of DAI in producing hybrid transformation animations that incorporate both texture and non-rigid transformations, we conduct qualitative experiments. Animations are generated using FAI, KVAI, and DAI, with the results shown in Fig. \ref{fig:DAI}. Using only FAI, the images fail to respond to non-rigid changes, while KVAI alone does not result in significant texture modifications. Our proposed DAI strategy successfully handles both texture and non-rigid changes, effectively achieving hybrid transformation animation.

\begin{table}[ht]
\centering
\captionsetup{font={small,stretch=1}, labelfont={bf}}
\caption{Quantitative ablation study Results.}
\resizebox{1.0\columnwidth}{!}{%
\renewcommand{\arraystretch}{0.95} 
\begin{tabular}{c||ccccc}
\hline
\multirow{2}{*}{\textbf{Method}} & \multicolumn{5}{c}{Metrics} \\
\cline{2-6}
 & \textbf{PIC} & \textbf{LPIPS$_T$} & \textbf{LPIPS$_M$} & \textbf{CLIP Score} & \textbf{PPL} \\
\hline
w/o FAI & 0.859 & 0.928 & 0.161 & 0.966 & 10.210 \\
w/o KVAI & 0.871 & 0.847 & 0.148 & 0.973 & 9.318 \\
w/o DAI & 0.877 & 0.804 & 0.130 & 0.972 & 8.847 \\
w/o ICA & 0.893 & 0.612 & 0.110 & 0.980 & 6.734 \\
Ours & \textbf{0.915} & \textbf{0.489} & \textbf{0.076} & \textbf{0.984} & \textbf{5.380} \\
\hline
\end{tabular}%
}
\label{tab:ablation}
\end{table}

\begin{table}[h]
\centering
\captionsetup{font={small,stretch=1}, labelfont={bf}}
\caption{Ablation study results on different subdatasets.}
\resizebox{\columnwidth}{!}{%
\renewcommand{\arraystretch}{1.0}
\begin{tabular}{c|c|ccccc}
\hline
\textbf{Subdataset} & \textbf{Method} & \textbf{PIC} & \textbf{LPIPS$_T$} & \textbf{LPIPS$_M$} & \textbf{CLIP Score} & \textbf{PPL} \\
\hline
\multirow{2}{*}{Material} & Ours & \textbf{0.940} & \textbf{0.230} & \textbf{0.028} & \textbf{0.989} & \textbf{2.525} \\
 & w/o FAI & 0.771 & 1.583 & 0.277 & 0.938 & 17.414 \\
\hline
\multirow{2}{*}{Non-rigid} & Ours & \textbf{0.947} & \textbf{0.431} & \textbf{0.059} & \textbf{0.990} & \textbf{4.736} \\
 & w/o KVAI & 0.820 & 1.453 & 0.264 & 0.958 & 15.987 \\
\hline
\multirow{2}{*}{Hybrid} & Ours & \textbf{0.848} & \textbf{0.860} & \textbf{0.154} & \textbf{0.970} & \textbf{9.462} \\
 & w/o DAI & 0.711 & 2.026 & 0.340 & 0.928 & 22.290 \\
\hline
\end{tabular}%
}
\label{ablation_2}
\end{table}

\section{Conclusion}
\label{sec:conclusion}

We introduce LASER, a tuning-free LLM-driven attention control framework that utilizes pre-trained text-to-image models to generate high-quality and smooth animations from multimodal inputs. Experimental results validate the superior performance of our method, consistently producing diverse and high-quality animations with enhanced detail and coherence. We believe our approach demonstrates significant potential and serves as an inspiration for continued advancements and future research in this field.

{
    \small
    \bibliographystyle{ieeenat_fullname}
    \bibliography{main}
}

\end{document}